\documentclass[sigconf]{acmart}

\AtBeginDocument{
  \providecommand\BibTeX{{
    \normalfont B\kern-0.5em{\scshape i\kern-0.25em b}\kern-0.8em\TeX}}}

\setcopyright{acmcopyright}
\copyrightyear{2022}
\acmYear{2022}
\acmDOI{10.1145/3534678.3539273}

\acmConference[KDD '22] {Proceedings of the 28th ACM SIGKDD Conference on Knowledge Discovery and Data Mining}{August 14--18, 2022}{Washington, DC, USA.}
\acmBooktitle{Proceedings of the 28th ACM SIGKDD Conference on Knowledge Discovery and Data Mining (KDD '22), August 14--18, 2022, Washington, DC, USA}
\acmPrice{15.00}
\acmISBN{978-1-4503-9385-0/22/08}

\usepackage{multirow}
\usepackage{amsmath}
\usepackage{hyperref}
\newtheorem{mydef}{Definition}
\begin{document}

\title{Continuous-Time and Multi-Level Graph Representation Learning for Origin-Destination Demand Prediction}



\author{Liangzhe Han}
\affiliation{%
  \institution{SKLSDE, Beihang University}
  \state{Beijing}
  \country{China}
  \postcode{100191}
}
\email{liangzhehan@buaa.edu.cn}

\author{Xiaojian Ma}
\affiliation{%
  \institution{SKLSDE, Beihang University}
  \state{Beijing}
  \country{China}
  \postcode{100191}
}
\email{xiaojianma@buaa.edu.cn}

\author{Leilei Sun}
\authornote{Also with Peng Cheng Laboratory, Shenzhen, China.}
\authornote{Corresponding Author.}
\affiliation{%
  \institution{SKLSDE, Beihang University}
  \state{Beijing}
  \country{China}
  \postcode{100191}
}
\email{leileisun@buaa.edu.cn}

\author{Bowen Du}
\authornotemark[1]
\affiliation{%
  \institution{SKLSDE, Beihang University}
  \state{Beijing}
  \country{China}
  \postcode{100191}
}
\email{dubowen@buaa.edu.cn}

\author{Yanjie Fu}
\affiliation{%
  \institution{University of Central Florida}
  \state{Florida}
  \country{USA}
  \postcode{32816}
}
\email{yanjie.fu@ucf.edu}

\author{Weifeng Lv}
\affiliation{%
  \institution{SKLSDE, Beihang University}
  \state{Beijing}
  \country{China}
  \postcode{100191}
}
\email{lwf@buaa.edu.cn}

\author{Hui Xiong}
\affiliation{%
  \institution{Hong Kong University of Science and Technology}
  \state{Hong Kong}
  \country{China}
  \postcode{999077}
}
\email{xionghui@ust.hk}

\renewcommand{\shortauthors}{Liangzhe Han, et al.}

\begin{abstract}
Traffic demand forecasting by deep neural networks has attracted widespread interest in both academia and industry society. 
Among them, the pairwise Origin-Destination (OD) demand prediction is a valuable but challenging problem due to several factors: (i) the large number of possible OD pairs, (ii) implicitness of spatial dependence, and (iii) complexity of traffic states. 
To address the above issues, this paper proposes a \textbf{C}ontinuous-time and \textbf{M}ulti-level dynamic graph representation learning method for \textbf{O}rigin-\textbf{D}estination demand prediction (\textbf{CMOD}).
Firstly, a continuous-time dynamic graph representation learning framework is constructed, which maintains a dynamic state vector for each traffic node (metro stations or taxi zones). 
The state vectors keep historical transaction information and are continuously updated according to the most recently happened transactions.
Secondly, a multi-level structure learning module is proposed to model the spatial dependency of station-level nodes.
It can not only exploit relations between nodes adaptively from data, but also share messages and representations via cluster-level and area-level virtual nodes.
Lastly, a cross-level fusion module is designed to integrate multi-level memories and generate comprehensive node representations for the final prediction.
Extensive experiments are conducted on two real-world datasets from Beijing Subway and New York Taxi, and the results demonstrate the superiority of our model against the state-of-the-art approaches.

\end{abstract}


\settopmatter{printacmref=True, printfolios=True}
\maketitle

\section{Introduction}
In recent years, deep learning techniques have been extensively extended into intelligent transportation systems.
Among all these applications, traffic prediction is the most attractive problem demonstrating its significance for urban construction, traffic controlling and route planning \cite{STGCN, DCRNN, ACGRN, MTGNN}.

Among existing work, most of them focus on forecasting how many people flow in or out an area.
Instead of merely forecasting the amount of passengers, Origin-Destination (OD) demand prediction also aims at predicting their destinations, which is rather important to understanding human mobility patterns.
Moreover, as there is a time series for each node pair, this is a distinctive and challenging problem due to much higher complexity than the most studied node-level prediction.
However, with the availability of large-scale transaction data, the problem attracts an increasing number of researchers to solve it.
Some recent studies split an area into regularly shaped grids and leverage Convolution Neural Networks (CNN) to capture spatial dependency of grids \cite{Contextualized, cityscale, AdversarialOD, MultiScale, multiresoluton}.
However, the demand is associated with stations or irregular traffic zones in many cases, which can not be solved with convolution filters.
Some studies generate representations for each OD pair, which can fit traffic graph topology but will significantly enlarge the complexity \cite{linegraph, zhang2021short, DNEAT}.
In addition, some studies focus on node representations making the complexity acceptable, while these methods are usually based on explicit but incomplete node relations \cite{GEML, MultiPerspective}.
\begin{figure}[!htbp]
    \centering
    \includegraphics[width=0.9\columnwidth]{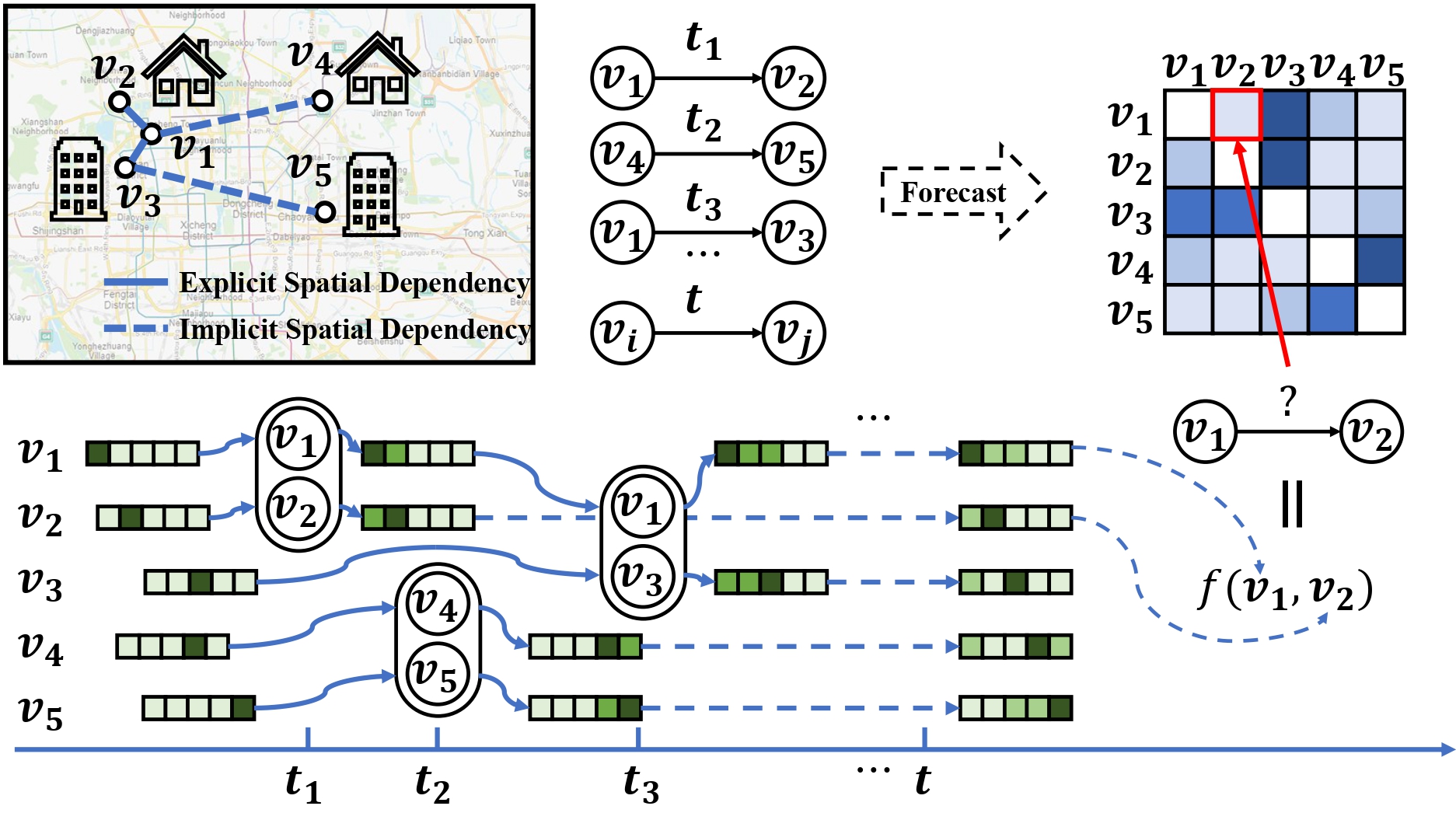}
    \caption{OD demand prediction with continuously evolving node representations.}
    \label{fig: motivation}
\end{figure}

Although there are some attempts on OD demand prediction, two important issues have rarely been discussed.
First, historical transactions are generally aggregated into demand snapshots, each of which contains demand in a fixed time window.
This operation will result in inevitable information loss.
Second, the spatial dependency in prior studies is always manually designed, which is intuitive but incomplete.
It has been demonstrated that there exists multiple types of relations in traffic \cite{multigraph}.
As shown in Figure \ref{fig: motivation}, the station in the west residential area is located close to other stations in the west.
But its demand can also be related to stations in the east residential area, as people from these stations may share a similar pattern going to central area.
However, hand designing is impossible to enumerate all potential relation types.
In summary, three challenges still remain:
(1) The time of transactions is a continuous feature which is unsatisfactory to process with fixed time windows.
It is challenging to model the complex temporal feature.
(2) There is implicit spatial dependency among traffic nodes. 
However, it is challenging to manually design optimal spatial dependency by hand.
(3) Each pair of nodes has its own time series. The quadratic amount of predicted values leads to the data sparsity problem. 

To address the above issues, this study proposes a novel \textbf{C}ontinuous-time and \textbf{M}ulti-level dynamic graph representation learning framework for \textbf{O}rigin-\textbf{D}estination demand prediction (\textbf{CMOD}).
The basic idea is shown in Figure \ref{fig: motivation}.
The framework maintains continuously updating node memories, which are vectors compressing and keeping historical transaction information to represent node status.
During each time of prediction, memories from last time are taken as input and updated according to the newly happened transactions.
Specially, the newly happened transactions are first leveraged to generate messages and update station-level node memories.
Next, we establish an adaptive multi-level structure by attention mechanism.
Then, the station-level messages are projected to cluster-level and area-level through their relations to update the corresponding memories.
Last, updated multi-level memories are fused for the final prediction, and they also act as the context of the next prediction.
Moreover, an objective function is designed to alleviate the data sparsity problem.
The main contributions are three folds:
\begin{itemize}
    \item \textit{A continuous-time dynamic graph representation learning framework is proposed for OD demand forecasting.}
    Different from the previous research, our method maintains continuous-time node representations and updated them continuously once a number of transactions are available. 
    As the evolutionary dynamics of stations could be learned in an extremely fine time scale, our method is promising to achieve higher prediction accuracy.
    \item  \textit{A hierarchical message passing module is proposed to model the spatial interactions of stations.}
    By sharing messages via the virtual cluster-level and area-level nodes, CMOD could exploit multi-level spatial dependence among stations.
    \item Extensive experiments have been conducted on two real-world datasets. The results not only demonstrate the superiority of our method over baselines, but also illustrate the ability of our method to capture the continuous evolving trajectory of station status.
\end{itemize}

What's more, the proposed method has potential to be further extended to other applications predicting features on edges, such as inter-country trade amount and network usage prediction.
The ideas to eliminate the effect of fixed time windows and exploit implicit node relations could also work in those scenarios.

\section{Preliminaries}
\begin{mydef}[Dynamic Transaction Graph]
Passenger demand from one location to another can be reflected in historical transaction records, such as taxi orders or metro records.
These records contain the origin, the destination and the departure time of passengers.
In this study, these transactions are organized as a continuous-time dynamic graph $\mathcal{G}=(\mathbb{V}, \mathbb{E})$,  where $\mathbb{V}=\{v_1, v_2, \cdots, v_N\}$ is a finite set of $N$ traffic nodes; $\mathbb{E}=\{e_1, e_2, \cdots, e_M\}$ is the set of $M$ timestamped transactions.
Each node represents a fixed station or a zone and an edge $e_m=(v_m^o, v_m^d, t_m)$ represents a passenger from $v_m^o$ to $v_m^d$ at time $t_m$.
Each node has a feature vector, and features of all nodes are denoted as $\mathbf{F}\in \mathbb{R}^{N\times d^F}$, where $d^F$ is the dimension of feature vectors. 
Moreover, the dynamic graph at a certain time $t$ is denoted as $\mathcal{G}_t=(\mathbb{V},\{e_k|t_k<t\})$, which contains all transactions before $t$.
\end{mydef}

\begin{mydef}[OD Demand Matrix]
OD demand matrix is a compressed format of passenger demand.
It contains the amount of demand between each pair of nodes in a period of time.
Formally, the OD demand matrix between $t$ and $t+\tau$ is denoted as $\mathbf{Y}^{t:t+\tau}\in \mathbb{R}^{N\times N}$.
The $(i,j)$-entry of $\mathbf{Y}^{t:t+\tau}$ represents how many passengers travel from $v_i$ to $v_j$ between $t$ and $t+\tau$: $\mathbf{Y}^{t:t+\tau}_{i,j}=|\{e_k|v_k^o = v_i \wedge v_k^d = v_j \wedge t\leq t_k< t+\tau\}|$, where $|\cdot|$ is the size of a set.
\end{mydef}

\begin{mydef}[OD Demand Prediction Problem]
Compared to demand of nodes, OD demand from node to node can better reveal human mobility.
OD demand prediction can not only estimate the amount of passengers in the future, but also gives an illustration about how they move, which is extremely helpful to transportation management.
Formally, given historical transaction records, the aim of OD demand prediction is to estimate OD demand matrix in the next period of time:
\begin{equation}\nonumber
    \hat{\mathbf{Y}}^{t:t+\tau} = f(\mathcal{G}_t, \mathbf{F}, \mathbb{W}),
\end{equation}where $\mathbb{W}$ is the set of learnable parameters. 
\end{mydef}

\begin{figure*}[!htbp]
  \centering
  \includegraphics[width=0.9\textwidth]{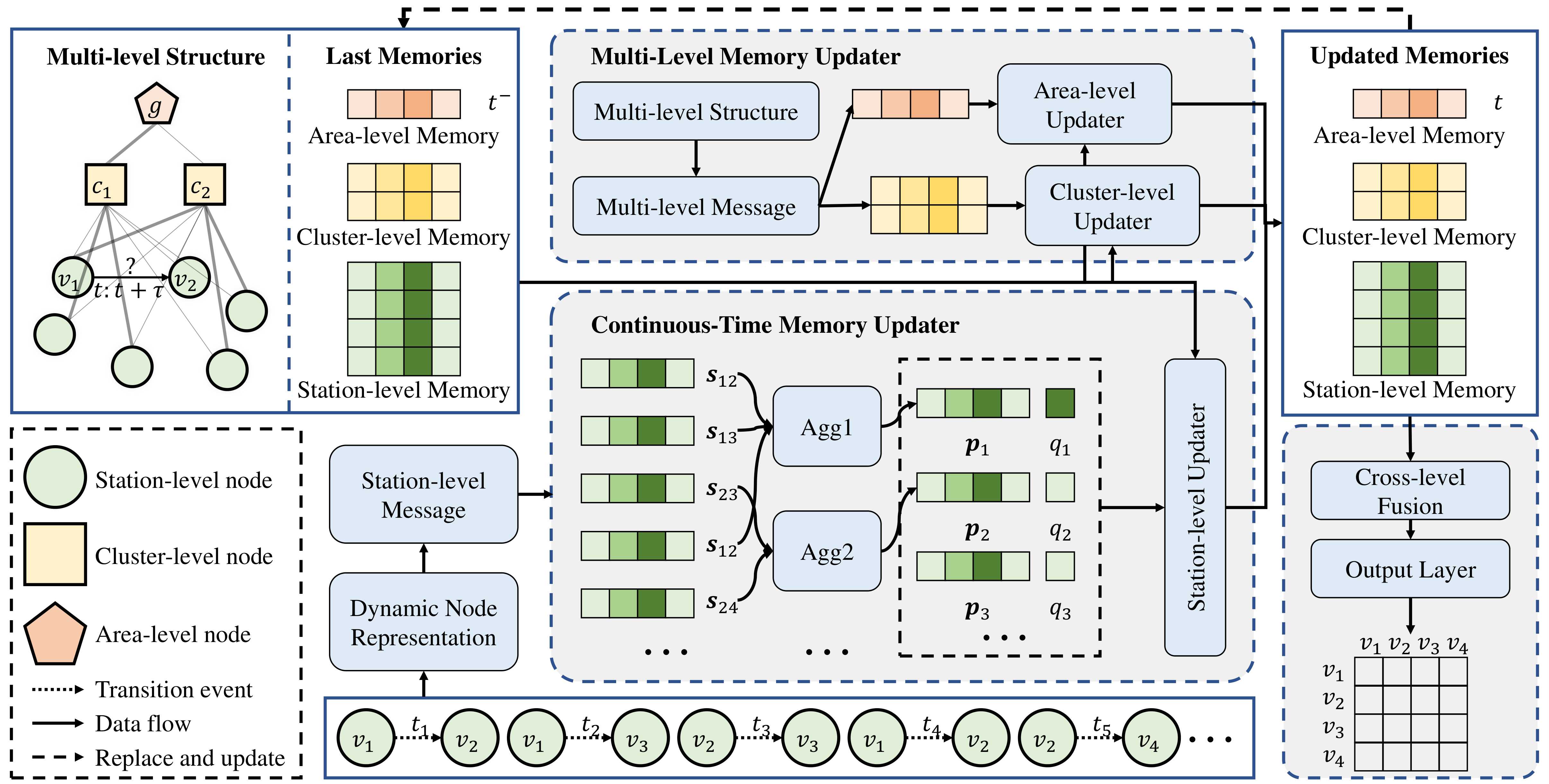}
  \caption{The overall framework of CMOD. It maintains continuously updated memory for each traffic node. When new transitions happen, the latest memories are used to represent these transitions. Then the representations are aggregated as messages for each node to update memories of corresponding station-level nodes. Meanwhile, a multi-level structure is established. Then the station-level messages are projected to cluster-level and area-level messages to update memory of nodes in these levels. Last, dynamic node representation is generated by cross-level fusion for the final prediction. 
  }
  \label{fig: archtecture}
\end{figure*}
\section{Methodology}
The overall framework of CMOD is shown in Figure \ref{fig: archtecture}.
The core idea of the framework is to maintain multi-level memories for nodes.
The multi-level structure is designed to capture spatial dependency between traffic nodes.
When transitions happen, this framework will update these memories with these streaming events.
And the updated memories, which compress all historical transactions and represent real-time node status, are utilized for the final prediction.
\subsection{Continuous-Time Node Representation}
To predict future OD demand, a general framework adopted by all previous work is based on discrete-time aggregation.
In another word, they all aggregate demand into demand snapshots with a fixed time window.
And sequence learning modules (e.g. GRU, LSTM) are leveraged to capture temporal dynamic on multiple demand snapshots.
One main drawback is that this process simply counts transactions in a time window and discards original continuous time information.
This will make it hard to capture important features.
For example, there are two time windows having similar amount of demand, but the first time window indicates demand increasing while another indicates demand decreasing.
However, simply counting transactions in a time window is incapable to distinguish them from each other.

Different from previous discrete-time OD demand prediction methods, CMOD is directly built on raw transition records, which views time information as continuous features and maintains continuous-time evolving representation for each traffic node (e.g. metro stations, taxi zones).
The raw records are represented as timestamped events, which are also streaming interactions between traffic nodes.
To get the node representations, there are two basic assumptions: 
1) the more recently an interaction happens, the more important it will be; 
2) the more frequently node i interact with node j, the more effect node j should have on node i. 
To this end, we expect representation of node i at time $t$ to be
\begin{equation}
  \mathbf{r}_i^t=\frac{\sum\limits_{(v_i, v_j, t^\prime)\in E^t}exp(-\lambda (t-t^\prime))\mathbf{r}_j^{t^\prime}}{\sum\limits_{(v_i, v_j, t^\prime)\in E^t}exp(-\lambda (t-t^\prime))},
  \label{node_rep}
\end{equation}
where $\mathbf{r}_i^t$ is representation of node $i$ at time $t$ and $\lambda$ is a hyperparameter which controls how fast the weights decay.
However, directly calculating node representations according to Equation \ref{node_rep} at time $t$ requires recomputing weights for all previous interaction when new transactions happen.
Here, we can choose to maintain two accumulators $\mathbf{a}$ and $b$ as node memory to update the representation in an online manner:
\begin{equation}
\begin{split}
  \mathbf{a}_i^t=exp(-\lambda (t-t^-))\mathbf{a}_i^{t^-}+\mathbf{r}_j^{t^-},\\
  b_i^t=exp(-\lambda (t-t^-))b_i^{t^-}+1,
  \end{split}
   \label{cal_b}
\end{equation}
where $\mathbf{a}_i^0=\mathbf{0}$, $b_i^0=1$ and $t^-$ is last update time of node $i$.
$\mathbf{a}_i^t$ is temporally weighted sum of neighbor node representations and acts as numerator in Equation \ref{node_rep}; and $b_i^t$ is temporal normalizer to acts as denominator in Equation \ref{node_rep}.
And node status at time $t$ is represented as
\begin{equation}
  \mathbf{r}_i^t=\frac{\mathbf{a}_i^t}{b_i^t}.
\end{equation}

The above procedure provides an inspiration to maintain dynamic node representations based on timestamped events.
However, in OD prediction, the amount of events is extremely huge, which makes it impossible to update node representation for each event.
Meanwhile, the procedure described above is lack of expressive power; it has no trainable parameters or meaningful initial features.
To address these issues, our dynamic node representation procedure is formally designed as following:
First, given a batch of newly happened events, we calculate an representation for each event; it combines node representation with some inherent features for a meaningful start:
\begin{equation}
  \mathbf{s}_k=[\mathbf{r}_j^{t^-};F_j], e_k=(v_i, v_j, t_k),
\end{equation}
where $t^-$ is last update time of node representations and $[\cdot;\cdot]$ is concatenation of two vectors.
Then for each node, we aggregate event representation involving it to get station-level messages, which are used to update station-level node status:
\begin{equation}
\begin{split}
  \mathbf{p}_{i}^t = \sum\limits_{(v_i, v_j, t_k)\in E^t-E^{t^-}}exp(-\lambda(t-t_k))\mathbf{s}_k,
\\
  q_{i}^t = \sum\limits_{(v_i, v_j, t_k)\in E^t-E^{t^-}}exp(-\lambda(t-t_k)).
  \end{split}
  \label{cal_q}
\end{equation}
Next, node memories, which consists of two accumulators similar as Equation \ref{cal_b}, are updated by the messages:
\begin{equation}
  \begin{split}
  \mathbf{a}_{i}^t = exp(-\lambda (t-t^-))\mathbf{a}_{i}^{t^-}+MLP(\mathbf{p}_{i}^t),\\
  b_{i}^t=exp(-\lambda (t-t^-))b_{i}^{t^-}+q_{i}^t.
  \end{split}
\end{equation}
And the updated node representations could be obtained as:
\begin{equation}
  \mathbf{r}_i^t=\frac{\mathbf{a}_{i}^t}{b_{i}^t}.
  \label{node_emb}
\end{equation}

Note that the message can also be extended with edge features such as user information and the transaction price, which is hard for previous snapshot-based methods.
\subsection{Multi-level Structure}
It is widely known that there exist multiple types of spatial dependency among nodes in traffic including geographical distance and functional similarity.
For example, in the morning peak, people come from different residual areas to a business area; moreover, adjacent subway stations covered by a big community may perform similarly.
However, the spatial dependency is hard to enumerate by hand.
Therefore, this study establishes an adaptive multi-level structure by the attention mechanism to automatically exploit the spatial dependency among nodes.
As shown in top-left of Figure \ref{fig: archtecture}, station-level nodes are aggregated to virtual cluster-level nodes and cluster-level nodes are aggregated to the virtual area-level node.
The rationale is that the attention mechanism can assign weights to determine how much a station belongs to a cluster and how much a cluster affects the global area status.
Station-level nodes that have similar OD demand patterns could be highly related to the same cluster and share useful information.

Formally, for the relations between stations and clusters, the relation matrix $\mathbf{A}^c_h\in \mathbf{R}^{N\times N^c}$ is computed as:
\begin{equation}
    \mathbf{A}^c_h=(\mathbf{W}^{c1}_h (\mathbf{r}^{t^-})^T)^T(\mathbf{W}^{c2}_h (\mathbf{r}^{c,t^-})^T), h=1,2,\cdots,H,
\end{equation}
where $\mathbf{r}^{t^-}\in \mathbf{R}^{N\times d}$ is station-level nodes representations, $\mathbf{r}^{c,t^-} \in \mathbf{R}^{N^c\times d}$ is cluster-level nodes representations, $\mathbf{W}_{\cdot}\in \mathbf{R}^{d^\prime\times d}$ is learnable parameters. 
Furthermore, multiple identical heads are leveraged here to capture relations in different aspects.
Meanwhile, relations between clusters and the whole area $\mathbf{A}^g_h\in \mathbf{R}^{N^c\times N^g}$ are computed similarly.
Here, $(i,j)$-entry of $\mathbf{A}^c_h$ represents the strength that node $i$ belongs to cluster $j$; similarly, $(i, 1)$-entry of $\mathbf{A}^g_h$ represents the strength that cluster $i$ affects the state of the whole area.
The first advantage of this multi-level structure is that unlike designing by hand, the attention mechanism is parameterized and can be optimized to obtain a proper clustering relation for OD demand prediction.
The second advantage is that based on dynamic representations of different levels, the spatial dependency can change adaptively in different situations.

\subsection{Multi-level Memory Updater}
\begin{figure}
  \centering
  \includegraphics[width=0.98\columnwidth]{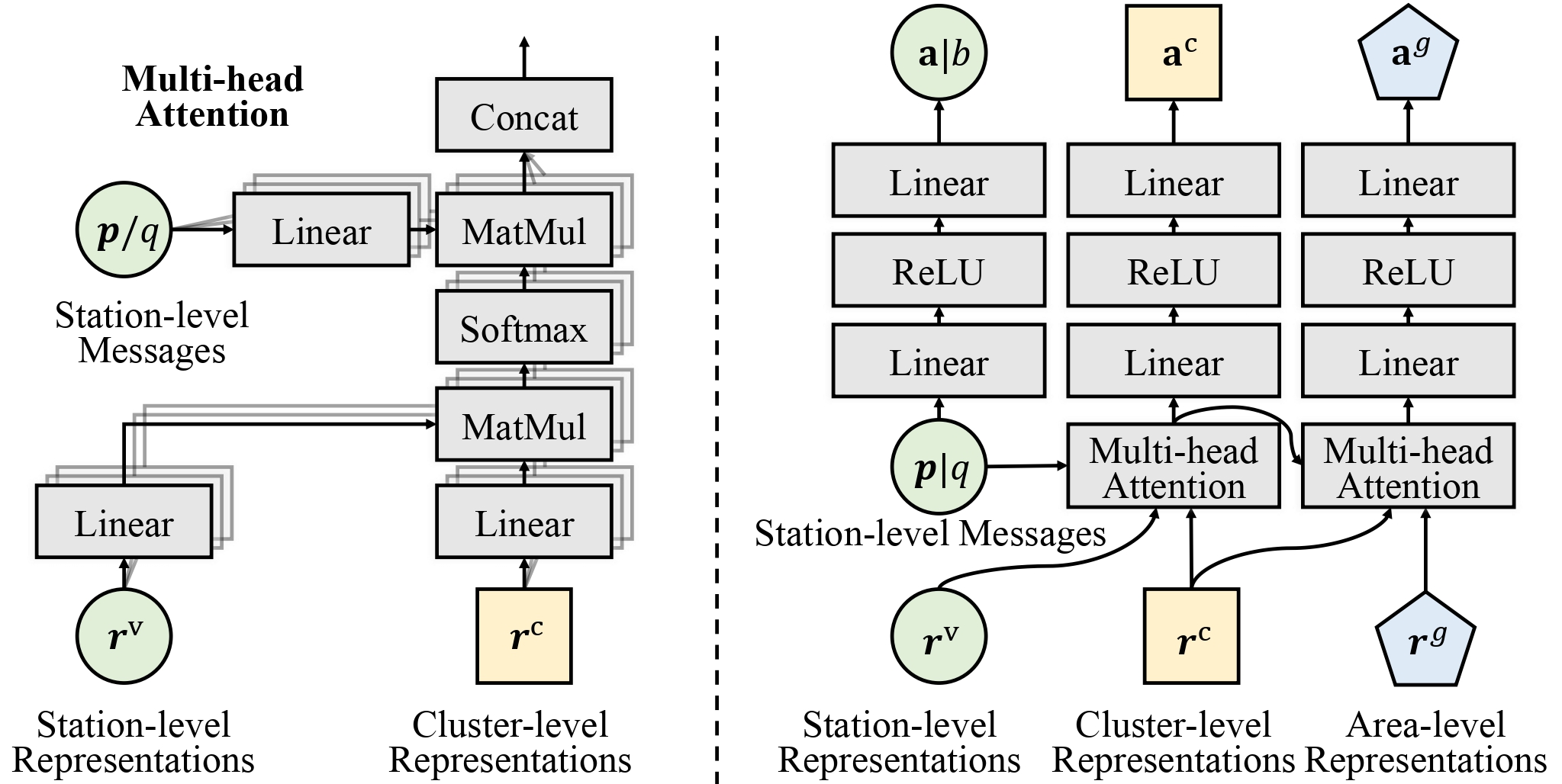}
  \caption{Illustration of multi-level memory updater.}
  \label{fig: multimseeage}
\end{figure}
Since we aim to maintain dynamic memories in three levels, it is essential to calculate corresponding messages to update them.
For example, if there are strong relations between a cluster-level node and several metro stations in business areas, its representation is supposed to be updated by transactions from these stations.
However, the raw transactions are only associated with station-level nodes.
Therefore, a module is proposed here to generate multi-level messages from station-level messages in Equation \ref{cal_q}. 

As shown in Figure \ref{fig: multimseeage}, with above relation matrices, messages are computed to update memories of multiple levels.
To be specific, messages for cluster $i$ are computed as:
\begin{equation}
\begin{split}
    \mathbf{A}^{c,m}_{h, j, i} = \frac{exp(\mathbf{A}^{c}_{h, j, i})}{\sum_{j=1}^Nexp(\mathbf{A}^{c}_{h, j, i})},\\
    \mathbf{p}^{c,t}_{i}=\bigg\|_{h=1}^{H}\sum_{j=1}^{N}\mathbf{A}^{c,m}_{h, j, i}(\mathbf{W}^{c3}_h\frac{\mathbf{p}_{i}^t}{q_{i}^t}),
    \end{split}
    \label{cluster_msg}
\end{equation}
where $\|$ is concatenation operation.
Similarly, messages for the graph memory are computed as:
\begin{equation}
\begin{split}
    \mathbf{A}^{g,m}_{h, i} = \frac{exp(\mathbf{A}^{g}_{h, i})}{\sum_{i=1}^{N_c}exp(\mathbf{A}^{g}_{h, i})},\\
    \mathbf{p}^{g,t}=\bigg\|_{h=1}^{H}\sum_{i=1}^{N^c}\mathbf{A}^{g,m}_{h, i}(\mathbf{W}^{g3}_h \mathbf{p}^{c,t}_{i}).
    \end{split}
\end{equation}
Then cluster-level and area-level node memories are updated:
\begin{equation}
  \mathbf{a}_{i}^{c,t} = exp(-\lambda (t-t^-))\mathbf{a}_{i}^{c,t^-}+MLP(\mathbf{p}^{c,t}_{i}),
\label{cluster_emb}
\end{equation}
\begin{equation}
 \mathbf{a}^{g,t} = exp(-\lambda (t-t^-))\mathbf{a}^{g,t^-}+MLP(\mathbf{p}^{g,t}).
\label{area_emb}
\end{equation}

In the above procedure, we discard normalizer $b$ for clusters and the area due to the fact that relations between levels are evolving all the time.
For example, in morning peak and evening peak, business cluster may weigh more to the whole area status, which makes it hard to aggregate the historical normalizer item.
Thus, we put the normalization procedure in Equation \ref{cluster_msg}.
And cluster-level and area-level node representations are directly set as their memories.

\subsection{Cross-level Fusion Module}
With previous modules, memories are updated to keep historical information for different levels.
To predict the OD demand matrix, updated node representations need to be extracted from those updated memories.
In the message generation part, the attention mechanism is utilized to model relations between different levels.
A stronger relation between a cluster and a station can not only indicate that the cluster should receive more messages from the station, but it also means the final station-level node representation should contain more information from the cluster.
For example, if one station belongs to a business cluster with a high weight, transactions involving the station should update memory of the cluster and memory of the cluster is also helpful to stations in it.

Since memories of different levels contain different parts of the information, a cross-level fusion module is proposed here to fuse memories from multiple levels.
Station-level node representations are calculated in Equation \ref{node_emb} and other-level representations are calculated in Equation \ref{cluster_emb} and \ref{area_emb}.
To fuse representations of clusters to station-level, relations between them are reutilized:
\begin{small}
\begin{equation}
    \mathbf{A}^{c,e}_{h, i, j} = \frac{exp(\mathbf{A}^{c}_{h, i, j})}{\sum_{j=1}^{N^c}exp(\mathbf{A}^{c}_{h, i, j})}, 
    \mathbf{r}_{i}^{c,t\prime} = \frac{1}{H}\sum\limits_{h=1}^{H}\sum\limits_{j=1}^{N^c}\mathbf{A}^{c,e}_{h, i, j}\mathbf{a}_j^{c,t}.
\end{equation}
\end{small}
The global area status are projected to station-level as:
\begin{small}
\begin{equation}
    \mathbf{A}^{g,e}_{h, i, j} = \frac{exp(\mathbf{A}^{g}_{h, i, j})}{\sum_{j=1}^{N^g}exp(\mathbf{A}^{g}_{h, i, j})}, 
    \mathbf{r}_{i}^{g,t\prime} = \frac{1}{H}\sum\limits_{h=1}^{H}\sum\limits_{j=1}^{N^c}\sum\limits_{k=1}^{N^g}\mathbf{A}^{g,e}_{h, i, j}\mathbf{A}^{g,e}_{h, j, k}\mathbf{a}_k^{g,t}.
\end{equation}
\end{small}
And the fused node representations are computed as following:
\begin{equation}
    \mathbf{Z}^t = [\mathbf{r}^{t};\mathbf{r}^{c,t\prime};\mathbf{r}^{g,t\prime}].
\end{equation}

How the attention mechanism captures the spatial dependency can be explained in two angles.
First, stations receive the same cluster information during cross-level fusion, which makes them partly similar.
Second, records involving a station send messages to clusters; when fusing representations for other stations, the fusion module makes them receive these messages through a bipartite-graph-like structure.
In both views, the more similar relations to clusters of two stations are, the more information can be shared.
\subsection{Output and Training}
The final prediction is then obtained based on the fused node representations.
For demand from node $i$ to node $j$, the prediction is calculated with concatenation of $\mathbf{Z}_{i}^t$ and $\mathbf{Z}_{j}^t$:
\begin{equation}
    \hat{\mathbf{Y}}_{i,j}^{t:t+\tau} = MLP(\mathbf{Z}_{i}^t;\mathbf{Z}_{j}^t).
\end{equation}

To predict OD demand for each pair of nodes, how to handle the situation that many pairs have no demand at a single time is important.
Here, a loss is customized for OD demand prediction.
The motivation is that more attention should be paid to non-zero demand and if one pair is unlikely to have demand, it is tolerable to predict a negative number.
And the loss is defined as:
\begin{equation}
    \begin{split}
        \mathcal{L}=\frac{1}{|\mathbf{Y}|}\sum\limits_{y\in \mathbf{Y}}((I_1(y)I_2(\hat{y}) + 1 - I_1(y))(y-\hat{y})^2),
        \\
        I_1(y)=\begin{cases}
            1, y=0\\
            0, y>0
        \end{cases},
        I_2(\hat{y})=\begin{cases}
            1, \hat{y}>0\\
            0, \hat{y}\le 0
        \end{cases}.
    \end{split}
\end{equation}
And the negative values in final prediction are replaced as zeros.
\section{Experiments}
\subsection{Datasets}
The performance of the proposed model is evaluated on two real-world datasets.
\textbf{BJSubway} contains transaction records generated in Beijing Subway from June to July in 2017.
\textbf{NYTaxi} contains taxi orders generated in Manhattan from January to June in 2019 \footnote{Data is avaiable at \href{https://www1.nyc.gov/site/tlc/about/tlc-trip-record-data.page}{https://www1.nyc.gov/site/tlc/about/tlc-trip-record-data.page}}.
More detailed statistic information of these datasets is shown in Table \ref{table: datasets}. 
Our code is available at \href{https://github.com/liangzhehan/CMOD}{https://github.com/liangzhehan/CMOD}.

\begin{table}[!h]
  \caption{Statistic information of datasets}
  \label {table: datasets}
  \begin{tabular}{c|cc}
   \toprule
   Dataset & BJSubway & NYTaxi\\
   \midrule
   \#Nodes & 268 & 63 \\
   \#Orders & 279,227,618 & 38,498,427 \\
   \#Train Days & 42 & 139 \\
   \#Validation Days & 7 & 21 \\
   \#Test Days & 7 & 21 \\
   Average Demand  & 2.1694 & 1.1164 \\
   Zero Order Ratio & 54.84\% & 66.15\%\\
   \bottomrule
  \end{tabular}
\end{table}
\begin{table*}
  \caption{Comparison results with baselines.}
  \label{table: Comparison}
  \begin{tabular}{cc|ccc|ccc}
    \toprule
    \multirow{2}{*}{Dataset} & \multirow{2}{*}{Method} & \multicolumn{3}{c|}{All OD Pairs} & \multicolumn{3}{c}{OD Pairs with Demand above Average} \\
    \cline{3-8}
    & & MAE $\downarrow$ & RMSE $\downarrow$ & PCC $\uparrow$ & MAE $\downarrow$ & RMSE $\downarrow$ & PCC $\uparrow$ \\
    \midrule
    \multirow{9}{*}{BJSubway} 
    & HA & 2.9003 & 8.1266 & 0 & 8.0378 & 18.3277  & 0 \\
    & LR & 1.9396 & 5.3547 & 0.7521 & 6.0566 & 11.7181 & 0.7322\\
    & XGBoost & 1.8048 & 5.7709 & 0.7040 & 5.9098 & 12.9627 & 0.6449\\
    \cline{2-8}
    & GEML & 1.7291±0.0123 & 4.6018±0.1138 & 0.8279±0.0075 & 5.3002±0.0982 & 10.1491±0.2983 & 0.8083±0.0086 \\
    & DNEAT & 1.4706±0.0099 & 5.7384±0.0311 & 0.7237±0.0033 & 5.4476±0.0365 &13.0661±0.0798  & 0.6488±0.0055 \\
    & TGN & 2.1031±0.1629 & 5.8927±0.5148 & 0.6755±0.0659 &  6.5592±0.3331& 13.0607±1.1772 & 0.6455±0.0781 \\
    & DyRep & - & - & - & - & - & -\\
    \cline{2-8}
    & CMOD & \textbf{1.4475±0.0202} & \textbf{3.6890±0.0319} & \textbf{0.8911±0.0020} & \textbf{4.5068±0.0437} & \textbf{8.1441±0.0697} & \textbf{0.8773±0.0021}\\
    \hline
    \multirow{9}{*}{NYTaxi} 
    & HA & 1.4593 & 2.6569 & 0 & 3.6041 & 5.7289 & 0 \\
    & LR & 0.6907 & 1.3611 & 0.8586 & 1.9939 & 2.8069 & 0.8164\\
    & XGBoost & 0.6881 & 1.3555 & 0.8599 & 1.9895 & 2.8052 & 0.8185\\
    \cline{2-8}
    & GEML & 0.6476±0.0033 & 1.3432±0.0093 & 0.8662±0.0015 & 1.8867±0.0138 & 2.7587±0.0198 & 0.8201±0.0025\\
    & DNEAT & 0.6495±0.0025 & 1.5179±0.0172 & 0.8252±0.0040 & 2.1922±0.0292 & 3.2834±0.0685 & 0.7581±0.0104 \\
    & TGN & 0.6516±0.0142 & 1.2947±0.0330 & 0.8747±0.0057 & 1.8387±0.0235 & 2.6435±0.0503  & 0.8311±0.0094 \\
    & DyRep & 0.6094±0.0032 & 1.2164±0.0089 & 0.8892±0.0013 & 1.7844±0.0296 & 2.5074±0.0398 & 0.8528±0.0017 \\
    \cline{2-8}
    & CMOD & \textbf{0.5926±0.0026} & \textbf{1.1795±0.0023} & \textbf{0.8959±0.0004} & \textbf{1.7244±0.0091} & \textbf{2.4263±0.0089} & \textbf{0.8618±0.0006} \\
    \bottomrule
  \end{tabular}
\end{table*}
\subsection{Baselines}
The detailed introduction for baselines are as following:
\begin{itemize}
    \item \textbf{HA} (Historical Average) computes historical average of OD demand matrix as prediction.
    \item \textbf{LR} (Linear Regression) is a regression model which exploits linear correlations between input and output. 
    \item \textbf{XGBoost} \cite{XGBoost} is a method based on gradient boosting tree. 
    \item \textbf{GEML} \cite{GEML} is an OD demand prediction model based on snapshots and pre-defined neighborhoods. Geographical neighborhood of GEML is defined by distance. 
    \item \textbf{DNEAT} \cite{DNEAT} is another OD demand prediction model based on snapshots and node-edge attention. The neighborhood definition of DNEAT is the same as GEML. 
    \item \textbf{TGN} \cite{tgn} is a dynamic graph representation learning model using graph attention network to obtain node representation. 
    Note that TGN is not originally designed for OD demand prediction, some modules limit their performance. 
    In comparison, its output module is set as the same as the proposed model. 
    One-hot encoding is used as node features. 
    \item \textbf{DyRep} \cite{DyRep} is a dynamic graph representation learning model using graph attention to aggregate neighbor messages. 
    One-hot encoding is also used as its node features. 
\end{itemize}

\subsection{Experiment Setups}
For both datasets, $\tau$ in prediction is set as 30 minutes.
The amount of cluster nodes $N_c$ is $\sqrt{N}$.
One-hot encoding is used as node features.
The proposed method is implemented with Pytorch toolbox on a machine with 4 Tesla T4 GPUs.
Adam optimizer with initial learning rate 0.0001 and early stopping strategy with patience 10 are utilized to train the proposed model in an end-to-end manner.
In dynamic representation part, the memory dimension is set as 256 and the representation dimension is set as 256.
In multi-level message part, the attention head number $H$ is set as 8 and the message dimension is set as 256.
The learning rate of all deep learning methods is chosen from [0.01, 0.001, 0.0001, 0.00001] according to the best performance on the validation set.
The best parameters on the validation set are selected to evaluate the performance.
All deep learning methods are repeated with different seeds for 5 times and the average value and the standard deviation are reported. 
Mean Average Error (MAE), Root Mean Square Error (RMSE) and Pearson Correlation Coefficient (PCC) are selected as metrics to compare.

\subsection{Comparison Results}
\begin{figure}
  \centering
  \includegraphics[width=\columnwidth]{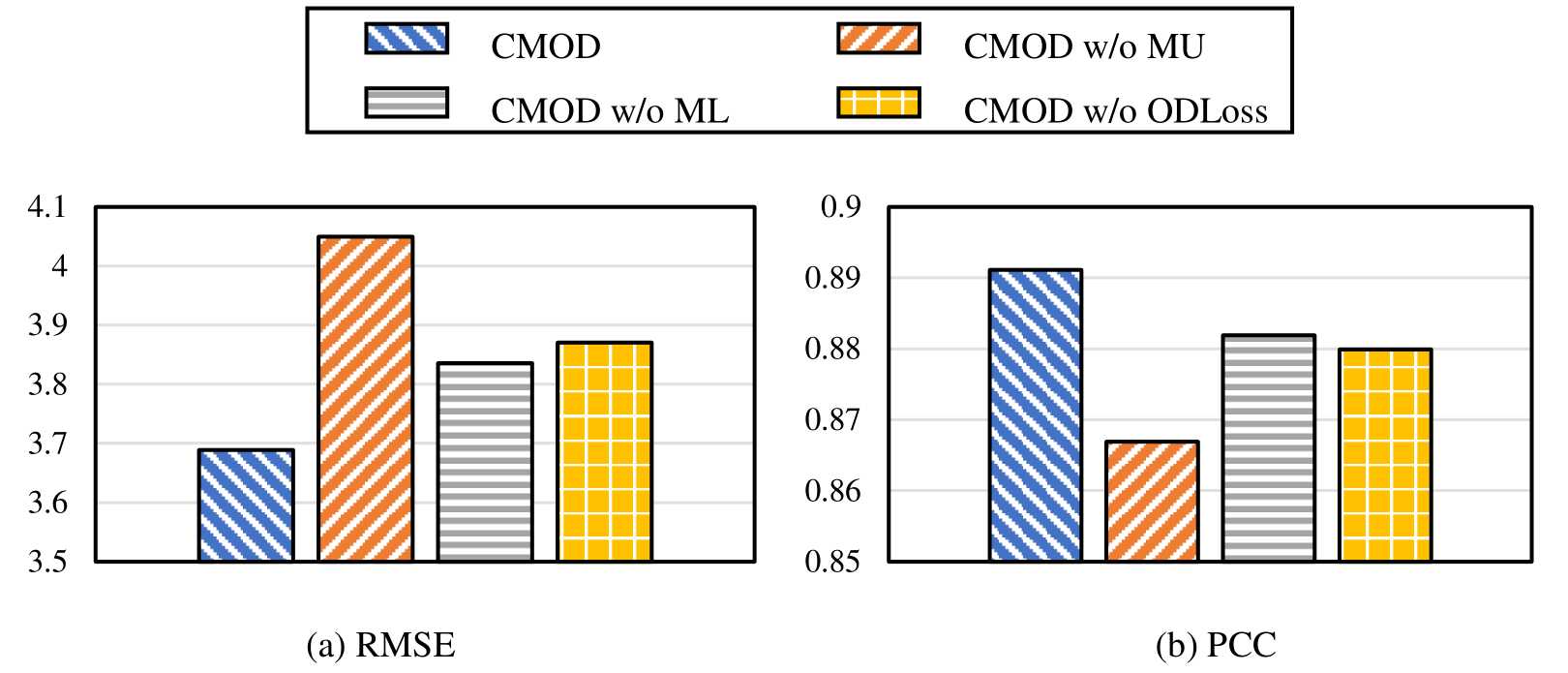}
  \caption{Ablation study.}
  \label{fig: Ablation}
\end{figure}
Table \ref{table: Comparison} summarizes the performance of all baselines. It can be observed:
\begin{figure*}
  \centering
  \includegraphics[width=0.9\textwidth]{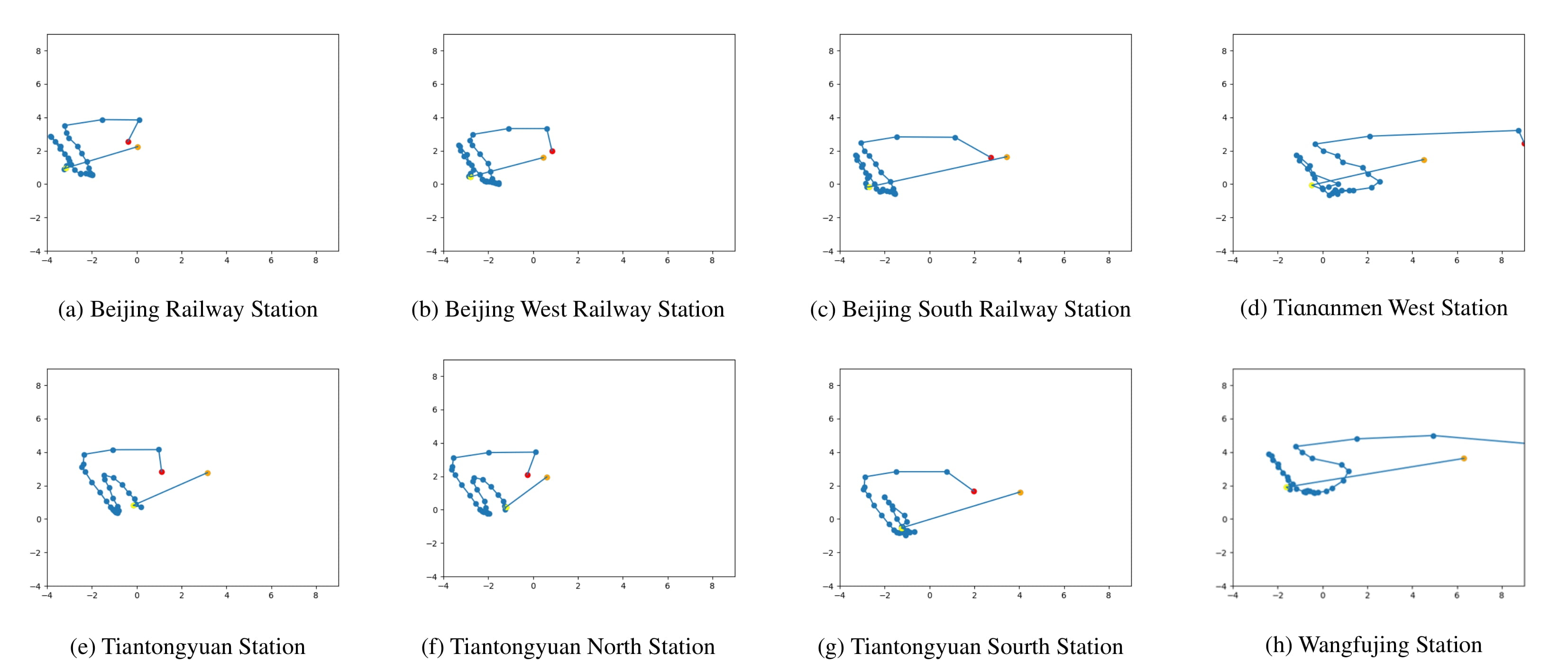}
  \caption{Illustration of evolving dynamic station representations.}
  \label{fig: rep}
\end{figure*}
(1) Overall, CMOD outperforms other methods in all cases, especially on BJSubway dataset, which suggests the effectiveness of our method to learn meaningful node representations for OD demand prediction.
(2) Though continuous-time dynamic graph representation learning methods (TGN and DyRep) are not originally designed for OD demand prediction, they perform better than other baselines on NYTaxi.
However, on BJSubway, TGN performs worse than other baselines and DyRep even encounters the out of memory problem.
One potential reason may be the intrinsic characteristic of OD demand prediction.
In this task, a large amount of edges (e.g., there are more than 200 million transactions in BJSubway) will make it hard for some designs to discover real demand patterns.
For example, TGN and DyRep both aggregate information from dozens of sampled neighbors.
In the context of dense edges, the sampling ratio is so small that brings more randomness and noises.
(3) Deep learning based methods perform better than simple statistic methods and traditional machine learning methods. 
The reason is that deep learning methods do not need designed features and have more expressive power to exploit complex and useful information from data.
One exception is that DNEAT only performs better on MAE.
The reason may be that its output module first predicts if there is demand as a probability from 0 to 1 and then multiples the probability with a predicted value as the final prediction.
This procedure will help when demand is low but will make it more unstable when demand is high.
Thus, compared to other deep learning methods, it performs better on MAE, where low-demand situations and high demand situations weigh the same, but performs worse on RMSE and PCC.

\subsection{Ablation Study}
\begin{figure}
  \centering
  \includegraphics[width=0.89\columnwidth]{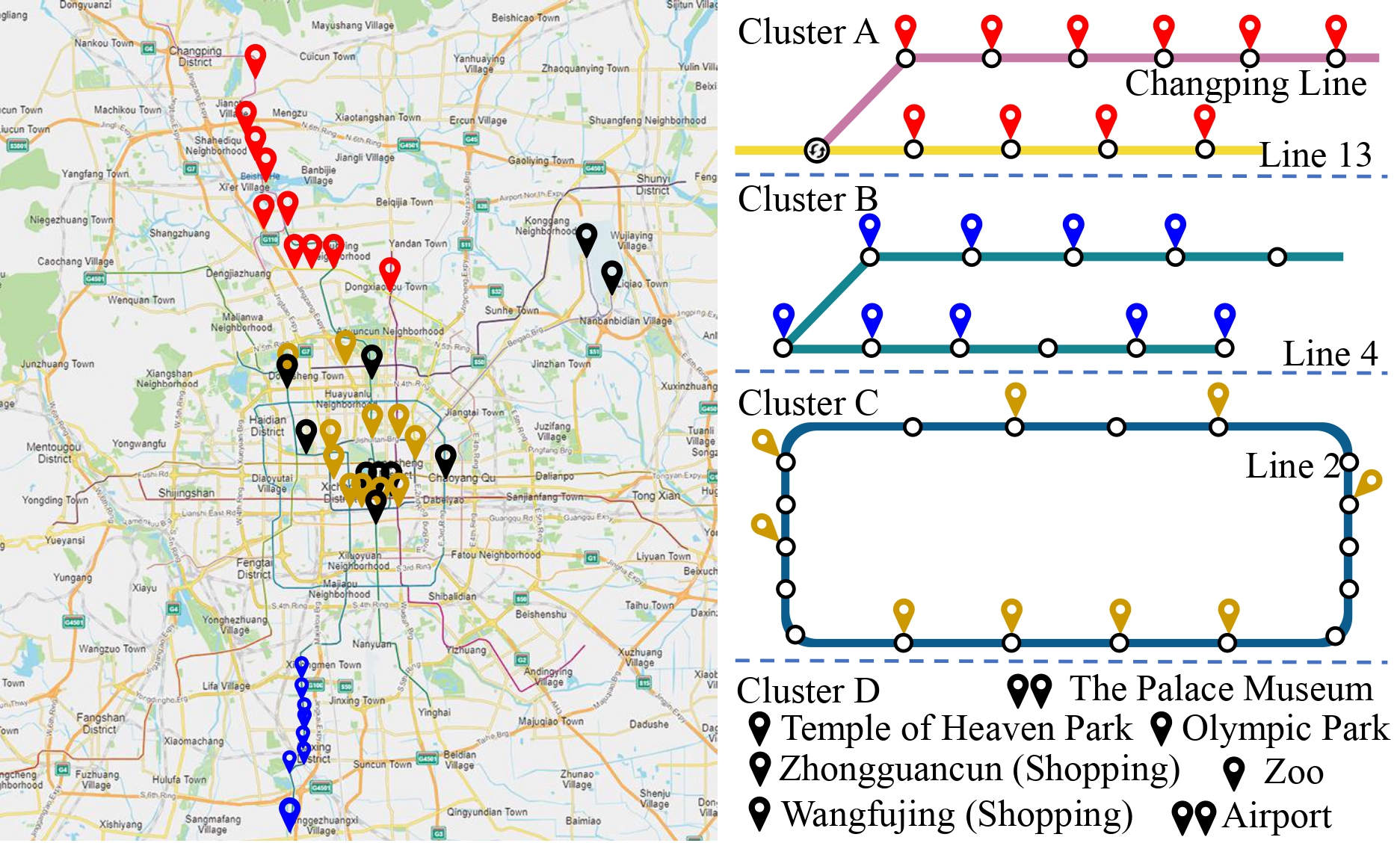}
  \caption{Interpretation of discovered clusters.}
  \label{fig: inter_attention}
\end{figure}
To demonstrate the effectiveness of each proposed component, an ablation study is conducted on BJSubway dataset with three variants:
\textbf{CMOD w/o ML} removes multi-level structure including memories of different levels, messages of different levels and cross-level fusion module.
\textbf{CMOD w/o MU} removes weighted memory updater and average messages and memories directly.
\textbf{CMOD w/o ODLoss} trains the model with MSE (Mean Square Error) loss.
The result is shown in Figure \ref{fig: Ablation}; it can be observed that:
(1) CMOD performs better than CMOD w/o ML; this demonstrates that the proposed attention-based multi-level structure can leverage spatial dependency for better prediction.
(2) The outperformance over CMOD w/o MU demonstrates that the memory updater which weigh transactions differently by time can provide valuable temporal information.
(3) After handling zeros by ODLoss, the model suits the OD demand prediction better than CMOD w/o ODLoss.
\begin{figure*}
  \centering
  \includegraphics[width=0.95\textwidth]{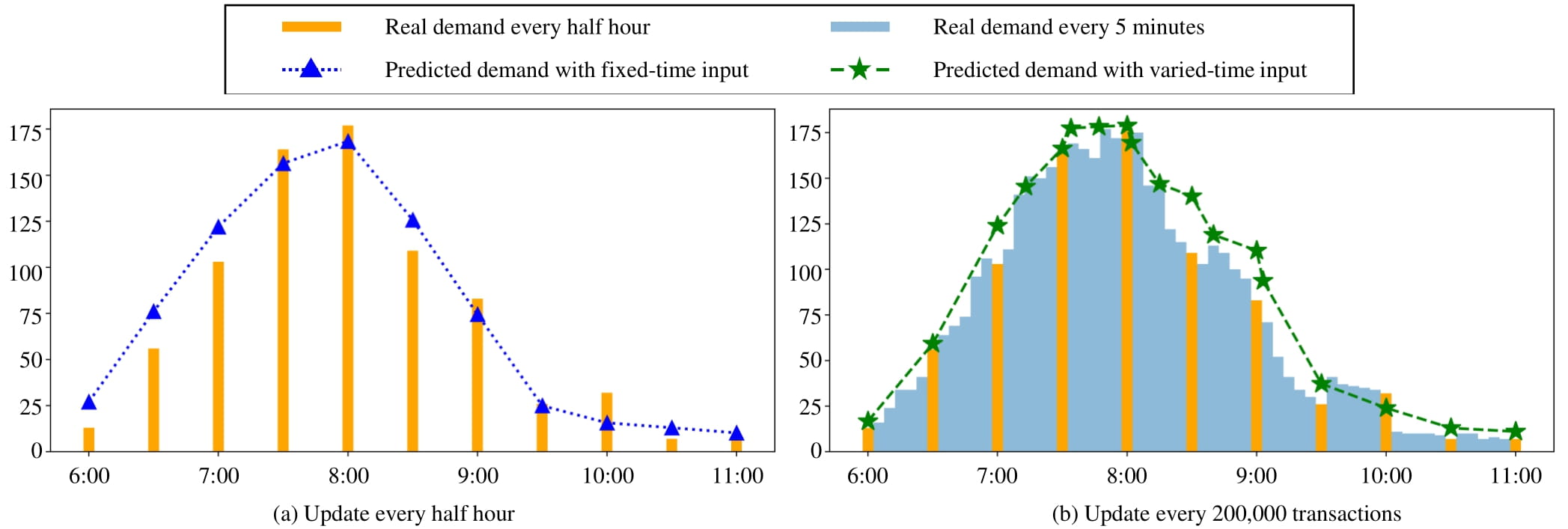}
  \caption{Prediction in two different settings.}
  \label{fig: varied}
\end{figure*}
\subsection{Illustration of Evolving Representations}
To avoid information loss during generating snapshots, the proposed CMOD is based on maintaining station representations.
Here, we conduct a case study on BJSubway to illustrate evolving patterns of station representations.
Specifically, representations of 8 stations from 6AM on first day to 6AM on second day are compressed to 2-dimensional vectors via Principal Component Analysis (PCA) and illustrated in Figure \ref{fig: rep}.
We represent station representations as points and connect them in chronological order.
The first point is marked as red and the final point is marked as orange.
From Figure \ref{fig: rep} (a, b, c), it can be observed that though metro stations at three railway stations in Beijing are nonadjacent, evolving patterns of their representations are similar.
This phenomenon demonstrates that the model discovers similar feature among them via transaction data.
Figure \ref{fig: rep} (d, h) show evolving patterns of Tiananmen West Station and Wangfujing Station.
These two stations are both famous tourist spots and their similar shape indicates their representations share another common evolving pattern.
It indicates that CMOD can automatically discover similar patterns from stations with similar functions.
Figure \ref{fig: rep} (e, f, g) are from representation of three metro stations around Tiantongyuan, the largest community in Asia.
It demonstrates that CMOD can automatically discover similar patterns from adjacent stations without predefined geographical information.
Thus, compared to designing by hand, the idea to establish relations from data is more powerful and helpful.

\subsection{Interpretation of Multi-level Structure}

To demonstrate how the multi-level structure works, another case study is further conducted on BJSubway.
To be specific, the highest weighted station-level nodes in $\mathbf{A}^c$ are selected to illustrate their locations in Figure \ref{fig: inter_attention}.
In Figure \ref{fig: inter_attention}, locations of nodes having highest weights with four clusters are illustrated on the map.
In cluster A (red marks), most of the stations are from two lines stretching out the center area; this is reasonable as people live in these places for a cheaper rent and have similar demands to travel to other working areas of the city.
In cluster B (blue marks) and cluster C (brown marks), most of the highly weighted stations are also distributed in a local area showing a similar pattern as cluster A.
In cluster D (black marks), the top weighted stations show another pattern; they don't gather in a small area.
Though these stations are nonadjacent directly, it is still reasonable.
These stations are all located around airports and tourist spots including zoo, the Palace Museum, Olympic Park and some shopping areas.
This indicates that cluster D discovers an implicit travelling pattern behind metro passenger data.
Thus, the results demonstrate that the multi-level structure can adaptively aggregate station-level nodes to clusters, establish relations among traffic nodes and benefit the final prediction.

\subsection{Prediction with Input of Varied Timespan}

In the above experiments, input of CMOD is transaction sets of every 30 minutes, which is for the comparison with previous OD demand prediction models.
Actually, the demand evolves continuously and it will be helpful to predict more frequently when it is in peak hour. 
Fortunately, another inherent advantage of CMOD is that unlike methods based on snapshots, the time granularity of input need not to set explicitly.
To demonstrate this, a case study is designed on BJSubway by varying the input timespan.
Specifically, if there are more than 200 thousand transactions during 30 minutes, they are split into several parts, each of which contains 200 thousand transactions at maximum.
As a result, memories are updated by different timespans.
As shown in Figure \ref{fig: varied} (a), if the memories are updated every 30 minutes, we can only obtain a sparse result.
However, when CMOD updates memories by varied timespans, it can obtain prediction result in Figure \ref{fig: varied} (b).
During 7:30 to 9:00, CMOD can update memories and predict OD demand more frequently.
It can be observed that the prediction with varied input timespan can fit denser real demand.
This demonstrates that CMOD is able to update memories with varied timespans and predict demand whenever the memories are updated. 

\section{Related Work}

\subsection{OD Demand Prediction}
Recently, there have been some attempts in three directions to introduce powerful deep learning into OD demand prediction problem.
In the first direction, researchers divided an area into regular-shaped grids and leveraged convolutional neural networks to capture their spatial dependency \cite{Contextualized, cityscale, AdversarialOD, MultiScale, multiresoluton}.
However, in many cases, OD demand are associated with irregular-shaped zones or stations on graph topology.
Although these methods could handle spatial dependency and temporal dynamics simultaneously, the requirement of grids limits their application.
Researchers in the second direction transferred edges to nodes in line graphs and leveraged methods for nodes to solve this problem \cite{linegraph, zhang2021short, DNEAT}.
They established relations based on underlying topology, while the complexity makes it difficult for a larger graph.
The third direction is to keep representations for nodes and reduced the complexity: researchers mainly utilized GCNs for each snapshot to capture spatial dependency and RNNs for multiple snapshots to capture temporal dynamics \cite{GEML, MultiPerspective}.
However, spatial dependency among traffic nodes was always designed by hand (e.g., geographical distance, last time demand); an improper relation may even hinder the prediction.
Moreover, to the best of our knowledge, all previous studies predicted OD demand based on snapshots, which will omit useful tendency information and bring ambiguity in choosing time granularity.

\subsection{Dynamic Graph Representation Learning}
Traditionally, research on graphs focused on static ones where nodes and edges are fixed \cite{GCN, GraphSage, GAT, GIN}.
They would fail for many applications involving dynamic graphs including streaming communication events in social media and streaming transactions in traffic.
Recent few years have witnessed a bunch of studies on dynamic graphs, and they fall into two categories: methods on discrete-time dynamic graphs (DTDGs) and methods on continuous-time dynamic graphs (CTDGs).
Methods on DTDGs represented a dynamic graph as multiple static graphs at different times; each static graph contains graph information in a period of time \cite{DynGEM, cc, DySAT, EvolveGCN}.
These studies took the temporal information into consideration but were still inflexible to fit more general cases.
Methods on CTDGs viewed edges in dynamic graphs as streaming timestamped events \cite{CTDNE, DyRep, JODIE, STREAMING, TGAT, tgn, TagGen}.
Most of them updated node representations after an event involving the node.
Although this architecture can handle temporal dependency and keep tendency information, these studies were lack of components to handle challenges in traffic including implicit spatial dependency and much denser events.

\section{Conclusion}
This study proposed a novel framework for the OD demand prediction problem.
The framework is based on modeling demand as a continuous-time dynamic graph.
It breakthroughs the limit of traditional snapshot-based models and can capture more useful information.
Moreover, a multi-level structure was established to adaptively exploit spatial dependency between traffic nodes.
Experiments on two real-world datasets demonstrated that the proposed model achieved the state-of-the-art performance. 
Case studies were also conducted to demonstrate the capability to handle input of arbitrary timespan and the effectiveness of the multi-level structure.
This work bonds promising CTDG methods and OD demand prediction for the first time, and it's also a new scenario for CTDG methods.
And the idea could also be further applied on more applications for edge-level prediction such as inter-country trade amount prediction and network usage prediction, where time information and node relations are also supposed to be handled carefully.
\begin{acks}
This work was supported by the National Natural Science Foundation of China (71901011, U1811463, 51991391, 51991395, U21A20516).
\end{acks}
\bibliographystyle{ACM-Reference-Format}
\bibliography{reference.bib}
\end{document}